\documentclass[11pt]{article}

\usepackage{arxiv}

\usepackage[utf8]{inputenc}
\usepackage[numbers]{natbib}
\usepackage[colorlinks=true,linkcolor=blue,citecolor=blue,urlcolor=blue]{hyperref}
\usepackage{amssymb}
\usepackage{amsmath}
\usepackage{booktabs}
\usepackage{subcaption}
\usepackage{enumitem}
\usepackage{graphicx}
\usepackage{xcolor}
\usepackage{textcomp}

\title{Scalable Deep Learning Framework for Global High-Resolution Land Use Reconstruction}

\author[1]{Amirpasha Mozaffari\thanks{Corresponding author: \texttt{amozafa@bsc.es}}}
\author[1]{Marina Casta\~{n}o}
\author[1]{Stefano Materia}
\author[1]{Etienne Tourigny}
\author[2]{Oscar Molina-Sedano}
\author[2]{Jordi Varela-Agrelo}
\author[2]{Dario Garcia-Gasulla}
\author[1]{Miguel Castrillo Melguizo}
\author[1]{Mario Acosta}
\author[1]{Amanda Duarte}

\affil[1]{Earth Science Department, Barcelona Supercomputing Center, Barcelona, 08034, Spain}
\affil[2]{AI Institute, Barcelona Supercomputing Center, Barcelona, 08034, Spain}

\date{}

\runninghead{Scalable Deep Learning for Global High-Resolution Land Use Reconstruction}

\begin{document}

\maketitle

\begin{abstract}
Uncertainty in the terrestrial carbon cycle remains a major constraint in climate projections, partly driven by the uncertainties affecting the land surface representation and variability in Earth system models. To address this limitation, we present a data-driven framework —AI4Land— for generating high-resolution historical reconstructions and future projections of key land surface variables. The framework follows a two-phase approach using a U-Net architecture. In the first phase, which is the focus of this work, it reconstructs annual land use and land cover by integrating coarse-resolution scenario data with static geophysical features. In a planned second phase, the resulting high-resolution maps will be used to predict dynamic biophysical variables, particularly leaf area index, at finer temporal scales. Trained on Earth observation data, the models learn to reproduce spatially explicit and physically consistent land surface patterns, extending temporal coverage to periods lacking direct observations. AI4Land was developed and trained on MareNostrum5, demonstrating how GPU-accelerated HPC infrastructure enables global-scale climate AI pipelines. The final product is a suite of open-source emulators designed for real-time coupling with digital twin platforms, such as those developed under the Destination Earth initiative. By delivering realistic and evolving land surface conditions on demand, this work aims to reduce critical uncertainties and improve the predictive power of next-generation climate simulations.
\end{abstract}


\section{Introduction}

The terrestrial carbon cycle remains a major source of uncertainty in climate projections \citep{booth2012high}, partly because land surface processes are not fully resolved. Land-use (LU) and land-cover (LC) changes are significant sources of uncertainty in estimating carbon fluxes \citep{gier2024representation}. Relying on coarse resolution and non-dynamic land boundary data can further lead to misrepresentation of land--atmosphere exchanges. Studies have shown that coarse land cover spatial resolutions can introduce substantial biases in simulated terrestrial carbon sequestration, affecting its magnitude, interannual variability, and spatial distribution \citep{zhao2010spatial}. By contrast, providing high-resolution ($\approx$1~km) land surface information, such as detailed land cover maps, leaf area index (LAI), or satellite vegetation indices like Normalized Difference Vegetation Index (NDVI), enhanced vegetation index (EVI), and soil-adjusted vegetation index (SAVI), allows climate models to capture fine-scale heterogeneity in vegetation and soil characteristics. This improved detail enhances the realism of carbon, water, and energy fluxes between the land and atmosphere. For example, using a new 1~km-resolution land parameter dataset in a land model produced pronounced spatial variability in soil moisture and surface fluxes (latent heat and radiation), whereas aggregating those inputs to $\sim$12~km led to a loss of about 31--54 \% of the spatial information \citep{li2024global}. High-resolution vegetation data are particularly crucial, since changes in LAI or greenness directly affect processes like photosynthesis (carbon uptake), evapotranspiration, and surface energy balance. For instance, a drop in LAI reduces canopy shading, increases ground net radiation, and dries out soil moisture \citep{boussetta2013impact, fang2019overview}. Accurate representation of high-resolution land boundary conditions in climate models is essential for reducing uncertainties in the terrestrial carbon cycle and for improving simulations of coupled carbon-water-energy fluxes \citep{gier2024representation}. Complementing this, advances in remote sensing of vegetation provide critical observational constraints that enhance the monitoring, understanding, and modelling of these land-atmosphere interactions \citep{prentice2024principles}.
Several datasets offer valuable insights into land surface boundaries, but each falls short in terms of spatial resolution, temporal (historical or future) and global coverage, or continuity \citep{hurtt2020harmonization, Copernicus_LAI_v2_1999_2020, winkler2020hilda+, luo20221, lv2025land, chen2020global}. Bridging this gap at a global scale requires not only advanced deep learning methodology but also GPU-accelerated HPC infrastructure to make training and inference computationally tractable. Building on these insights and datasets, our work aims to reduce uncertainties in terrestrial carbon cycle representation by developing high-resolution land surface boundary datasets that span historical, contemporary, and future periods. By integrating satellite-era observations with reconstructions for pre-observation times and projections for observation-limited futures, we provide temporally continuous, spatially detailed datasets for use in climate and weather models. These datasets were produced using EuroHPC compute allocations on MareNostrum5 at BSC and are designed to improve the representation of land surface heterogeneity, enabling more accurate simulations of carbon, water, and energy fluxes over time.

\section{Methods}

To generate temporally continuous, spatially detailed land surface boundaries, we focus in this work on reconstructing the slow-varying land surface boundary condition LU/LC at high spatial resolution; a planned second phase will extend the framework to dynamic biophysical variables such as LAI.

We have generated historical and future annual LU at high spatial resolution (1~km), derived from coarse-resolution inputs (0.25\textdegree, $\sim$28~km). The goal is to produce high-resolution LU maps for a target year $t$ ($H_t$), with each pixel classified into predefined categories (e.g., Forest, Cropland, Urban). Ground-truth labels are sourced from the high-resolution HILDA+ LU/LC dataset~\citep{winkler2020hilda+}. While the original HILDA+ classification scheme contains 13 classes, we consolidated these into 8 distinct LU categories to reduce class imbalance and streamline the learning objectives. The input tensor for year $t$ is a multichannel stack comprising Land-Use Harmonization~2 (LUH2) data for year $t$~\citep{hurtt2020harmonization} providing 12 fractional LU variables and two land surface parameters standardized to unit variance on a $0.25^{\circ} \times 0.25^{\circ}$ ($\sim$31~km) grid, static topographic features (elevation, slope, and aspect)~\citep{GEBCO2024}, high-resolution soil characteristics including clay, sand, and organic content from PNNL global datasets \citep{li2024global}, and an autoregressive prior from year $t-1$ or $t+1$ to enable bidirectional prediction.

\subsection{Preprocessing}

To ensure consistency across diverse data sources, we
established a standardized preprocessing pipeline relying on the
Climate Data Operators (CDO) software \citep{schulzweida2019cdo}. All inputs were remapped using nearest-neighbour interpolation to align with the HILDA+ WGS84 grid at 1~km resolution. For soil data, a non-linear depth-weighted average ($w = [0.40, 0.25, 0.15, 0.10, 0.05, 0.05]$) was applied to prioritize root-zone interactions. The final processed tensors were stored in Analysis-Ready Cloud Optimized (ARCO) Zarr format \citep{newman2024zarr}, chunked in $512 \times 512$-pixel blocks for efficient parallel I/O during distributed training.

\subsection{Model Architecture}

We implemented the LU reconstruction and projection using a U-Net architecture \citep{ronneberger2015u} for $512 \times 512$-pixel samples. Since all inputs are aligned to a WGS84 longitude--latitude grid at 1~km resolution, each patch covers approximately $512 \times 512$~km at the equator, and per-pixel classification is performed independently via dense semantic segmentation. The model employed a standard U-Net with 35 base channels, accepting inputs from 14 LUH2 variables across 2 time steps, 6 static features, and 1 HILDA prior. Its encoder consisted of four max-pooling and double-convolution blocks, mirrored by a symmetric decoder with upsampling and skip connections, terminating in a $1 \times 1$ convolution for final segmentation. To encode the categorical target, we utilized entity embeddings \citep{guo2016entityembeddingscategoricalvariables}. During training, 60\% of the autoregressive prior is randomly masked to prevent the model from trivially copying the high-resolution prior and force it to learn the mapping from the coarse LUH2 inputs and static features, which is essential for inference in periods without ground truth. As we produce both historical and future projections, we have trained two separate models: one for historical data (using LUH2h) and one for future projections (using LUH2f).

\begin{figure}[htbp]
  \centering
  \begin{subfigure}[b]{0.4\textwidth}
    \centering
    \includegraphics[width=\linewidth]{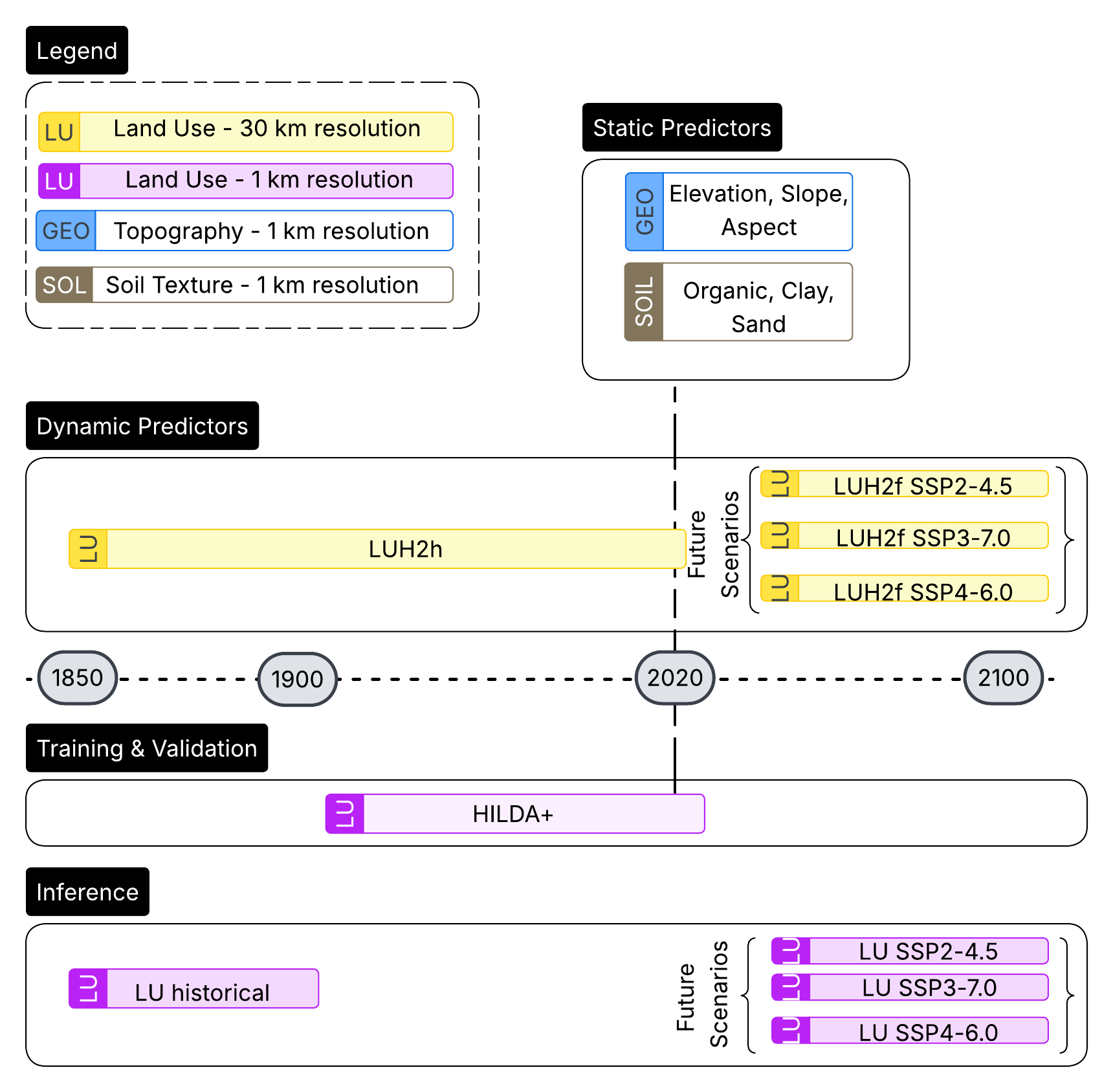}
    \caption{Spatio-Temporal Data Scope}
    \label{fig:method_data}
  \end{subfigure}
  \hfill
  \begin{subfigure}[b]{0.50\textwidth}
    \centering
    \includegraphics[width=\linewidth]{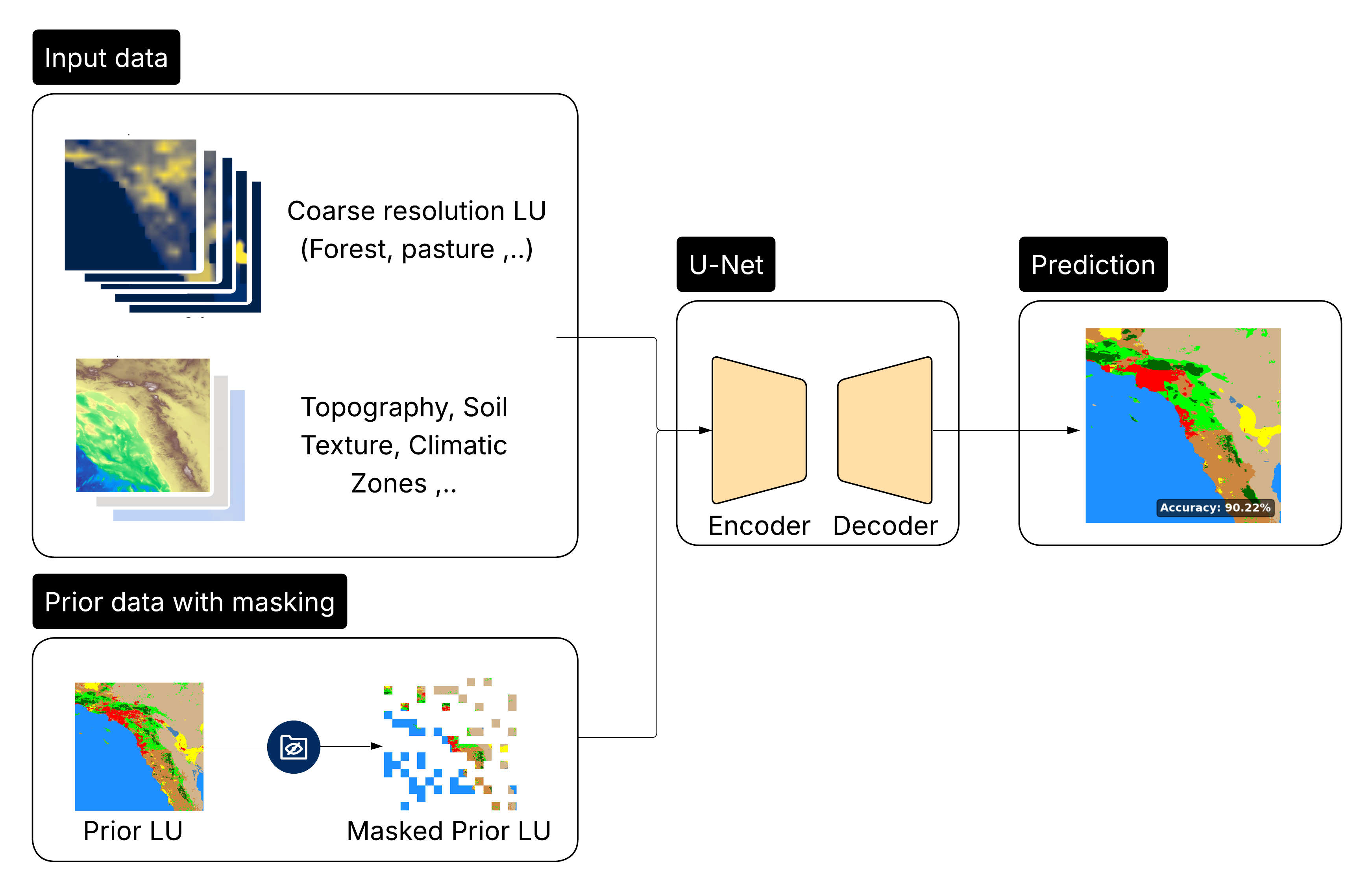}
    \caption{Reconstruction Pipeline}
    \label{fig:method_arch}
  \end{subfigure}
  \caption{The AI4Land Framework. (a) Data Availability: The timeline illustrates the critical resolution gap: while coarse dynamic forcing (LUH2) covers the full centennial scope (1850--2100), high-resolution ground truth (HILDA+) is limited to the satellite era. We fuse these dynamic inputs with high-resolution static priors to reconstruct the missing historical and future boundaries. (b) Reconstruction Pipeline: The U-Net backbone fuses heterogeneous inputs: coarse dynamic variables are upsampled and concatenated with fine-scale static features. To enforce temporal consistency, a randomly masked land use map from the adjacent time-step ($t \pm 1$) is injected as an autoregressive prior.}
  \label{fig:combined_method}
\end{figure}

\subsection{Distributed Training on MareNostrum5}


We trained the model on MareNostrum5 using Distributed Data Parallelism (DDP) via the Hugging Face Accelerate library~\citep{accelerate}, scaling from a single node (4 NVIDIA H100 GPUs) up to 8 nodes (32 H100 GPUs). In each epoch, we trained on 10,000 batches of 8 samples and validated on 2,000 batches of 8
samples, with the training dataset comprising 800,000 carefully selected samples. Each epoch on a single node required approximately 2 hours and 40 minutes, including training, validation, and I/O overhead.

\begin{figure}[htbp]
  \centering
  \includegraphics[width=0.6\linewidth]{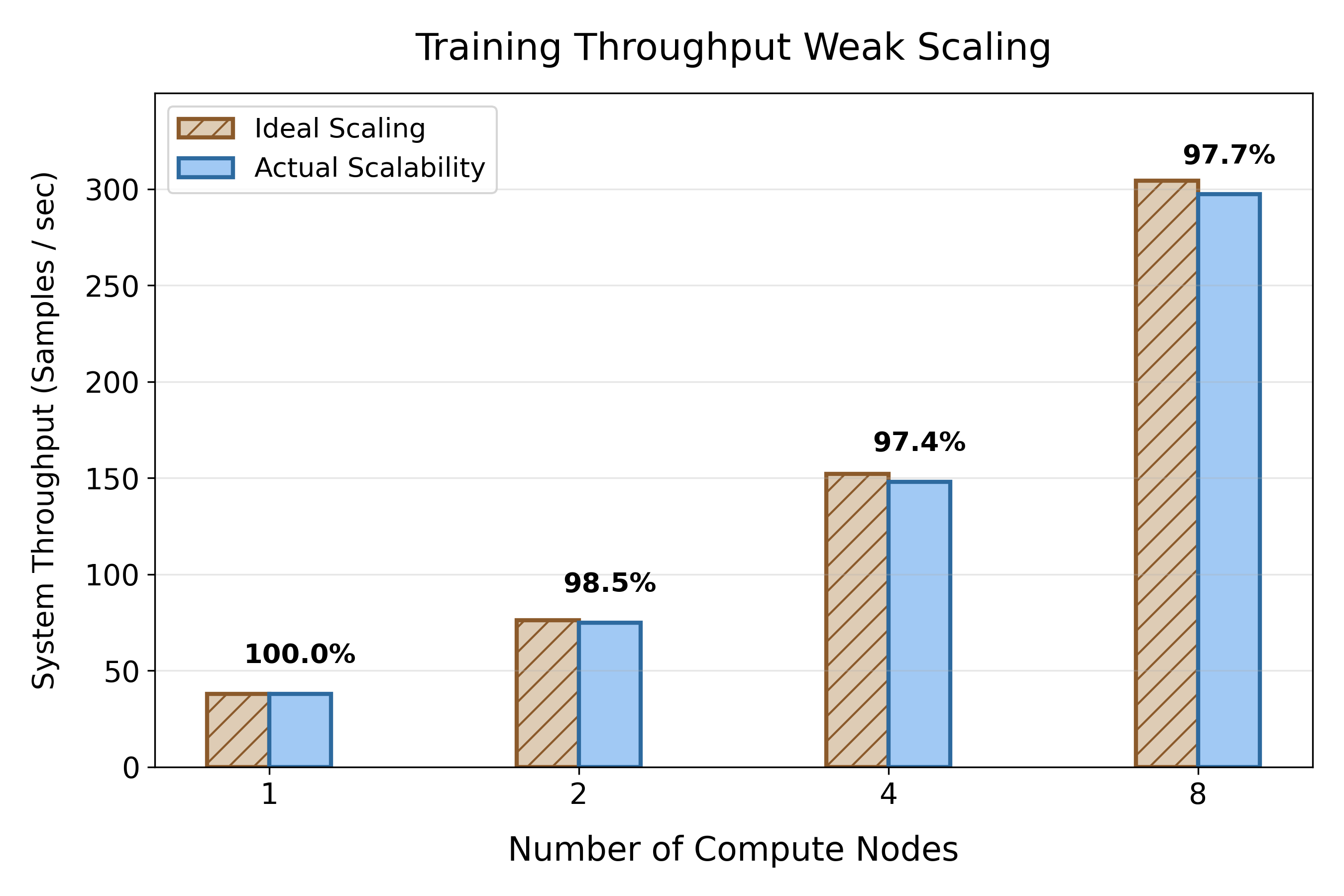}
  \caption{Weak scaling throughput of the AI4Land DDP training pipeline on
  MareNostrum5. System throughput (samples/sec) is shown for 1, 2, 4, and
  8 compute nodes alongside ideal linear scaling. Parallel efficiency remains
  above 97\% up to 8 nodes (32 H100 GPUs), demonstrating near-linear
  scalability of the Hugging Face Accelerate-based distributed training.}
  \label{fig:throughput}
\end{figure}

To quantify the scalability of the pipeline, we conducted a weak scaling analysis in which the per-node workload was kept constant while the number of nodes was increased from 1 to 8. As shown in Figure~\ref{fig:throughput}, the DDP training pipeline achieves near-linear scaling throughout this range, retaining 98.5\%, 97.4\%, and 97.7\% of ideal throughput at 2, 4, and 8 nodes, respectively. At maximum scale (8 nodes, 32 H100 GPUs), the system sustains approximately 300 samples per second, confirming that inter-node communication overhead on MareNostrum5 remains negligible.

\subsection{Evaluation}

We employed a two-phase evaluation strategy combining Grid-Based Partitioning and Farthest Point Sampling (FPS) to eliminate spatial autocorrelation leakage between training and testing sets. The spatial domain is divided into a coarse grid of $30 \times 30$ bins, with entire grid cells assigned exclusively to training, validation, or testing sets. Within these regions, FPS selects maximally distant points to ensure uniform spatial coverage. From approximately 214 million land pixels, we subsampled a final dataset of 448,000 points, stratified into 320,000 training, 64,000 validation, and 64,000 testing samples. To prevent data leakage, we applied a temporal split: years 1960–2000 were used for training and 2001–2015 for testing.

\begin{figure}[htbp]
  \centering
  \includegraphics[width=0.7\linewidth]{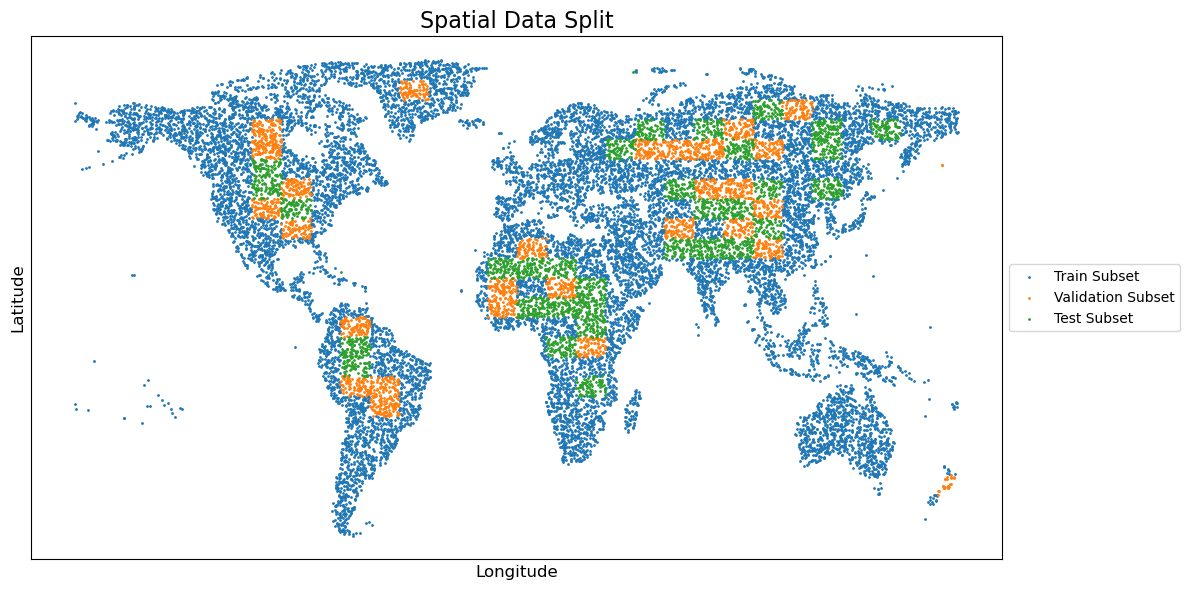}
  \caption{Spatial distribution of the training, validation, and test sets. Grid-based partitioning assigns entire macro-grid cells exclusively to each split, ensuring strict physical separation and eliminating spatial autocorrelation leakage between training and testing sets.}
  \label{fig:data_split}
\end{figure}

\subsection{Inference}

To generate spatially consistent global land use maps, the system employs a distributed inference pipeline processing the globe in overlapping $512 \times 512$ patches. To generate seamless global maps, we employ a sliding window approach with a Gaussian weighting function $W$ to merge overlapping predictions. The final probability $\bar{\mathbf{P}}(x, y)$ is computed by normalizing the weighted sum of  overlapping patches $\Omega_{(x,y)}$:

\begin{equation}
    \bar{\mathbf{P}}(x, y) = \frac{\sum_{k \in \Omega}
    \mathbf{p}^{(k)}(x, y) \cdot W(u_k, v_k)}{N(x, y)}
\end{equation}

The final class is derived via $\arg\max_{c}$, ensuring
smooth transitions across patch boundaries without blocky or sharp border artifacts.

\section{Results}

\subsection{Model Performance}

The U-Net architecture, a cornerstone of our LU classification
methodology, was rigorously trained on a dataset comprising 800,000 carefully selected samples. As we have to produce both historical and future projections, we have tackled this effort with two models: one for predicting historical data and one for predicting future data. Since the performance metrics for the historical projection model and the future projection model are nearly identical, we use the average performance metrics for both models. Following this extensive training phase, the model's performance was assessed, yielding a mean Intersection over Union (mIoU)---the average across classes of the intersection-over-union between predicted and ground-truth pixel sets---of 0.805. This metric corresponds to an overall classification accuracy of
94.67\%, achieved under a specific configuration that involved a 60\% masking strategy during the evaluation.

\begin{figure}[htbp]
  \centering
  \includegraphics[width=0.7\linewidth]{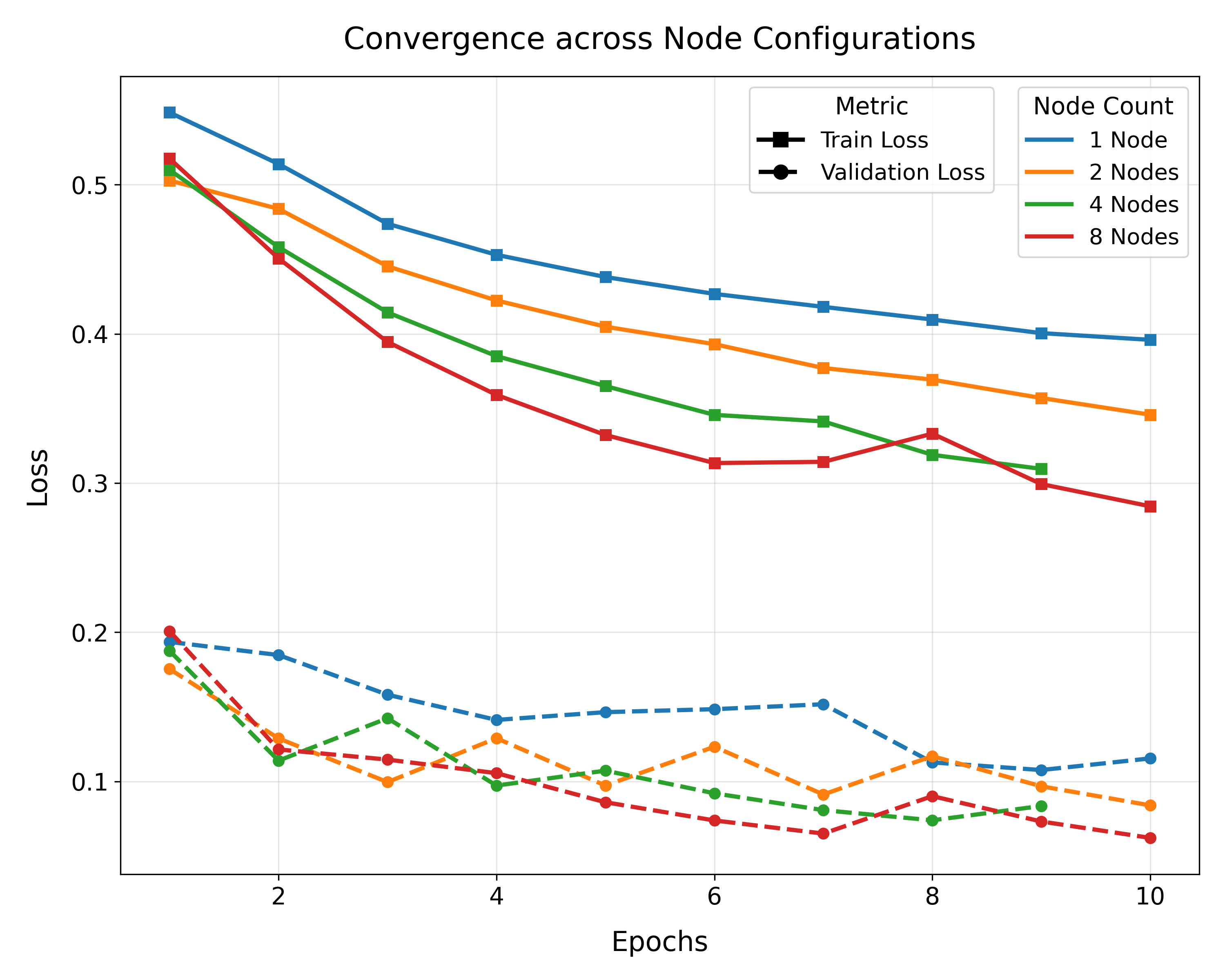}
  \caption{Convergence across node configurations on MareNostrum5. Solid lines with square markers denote training loss; dashed lines with circle markers denote validation loss, for 1, 2, 4, and 8 nodes. Configurations with more nodes consistently reach lower loss at equivalent epoch counts, with the 8-node run (32 H100 GPUs) achieving a final training loss of ${\approx}$0.29 versus ${\approx}$0.40 for the single-node baseline.}
  \label{fig:training_curves}
\end{figure}

Figure~\ref{fig:training_curves} shows training and validation loss across epochs for all node configurations (1, 2, 4, and 8 nodes). In all cases the training loss remains substantially higher than the validation loss, which is explained by the use of input masking during training whereas validation is performed on unmasked inputs. Notably, increasing the node count improves convergence per epoch: after 10 epochs the 8-node configuration achieves a training loss of approximately 0.29, compared to 0.40 for the single-node baseline, with validation loss similarly improving from ${\approx}$0.12 to below 0.10. This behaviour is consistent with the larger effective batch size under multi-node DDP, which provides a better gradient estimate per update step. Combined with the near-linear throughput scaling shown in Figure~\ref{fig:throughput}, this confirms that multi-node training on MareNostrum5 delivers both faster and better-converged solutions.

There is a performance disparity that was dramatically amplified for those LU types that were underrepresented in the training dataset. For these minority classes, the model's ability to generalize and accurately predict decreased substantially, resulting in a notable drop in the IoU, in the urban class, to 0.46. This outcome underscores the model's sensitivity to class imbalance and highlights the need for further strategies to improve generalization across all LU types, particularly those with fewer training examples. Table~\ref{tab:model_results}
shows the accuracy (\%), F1 score, and IoU across the classes.

\begin{table}[ht]
    \centering
    \caption{\textbf{Global Evaluation Results.} Per-class segmentation
    performance of the AI4Land U-Net model. Average loss: 0.1434.}
    \label{tab:model_results}
    \begin{tabular}{lccc}
        \toprule
        \textbf{Class} & \textbf{Accuracy (\%)} & \textbf{F1} &
        \textbf{IoU} \\
        \midrule
        Urban           & 48.68 & 0.632 & 0.463 \\
        Cropland        & 90.62 & 0.897 & 0.815 \\
        Pasture         & 89.52 & 0.897 & 0.814 \\
        Forest          & 94.93 & 0.941 & 0.889 \\
        Grass/shrubland & 81.47 & 0.833 & 0.715 \\
        Other land      & 97.05 & 0.971 & 0.944 \\
        Water           & 99.24 & 0.995 & 0.991 \\
        \bottomrule
    \end{tabular}
\end{table}

\begin{figure}[htbp]
  \centering
  \includegraphics[width=0.90\linewidth]{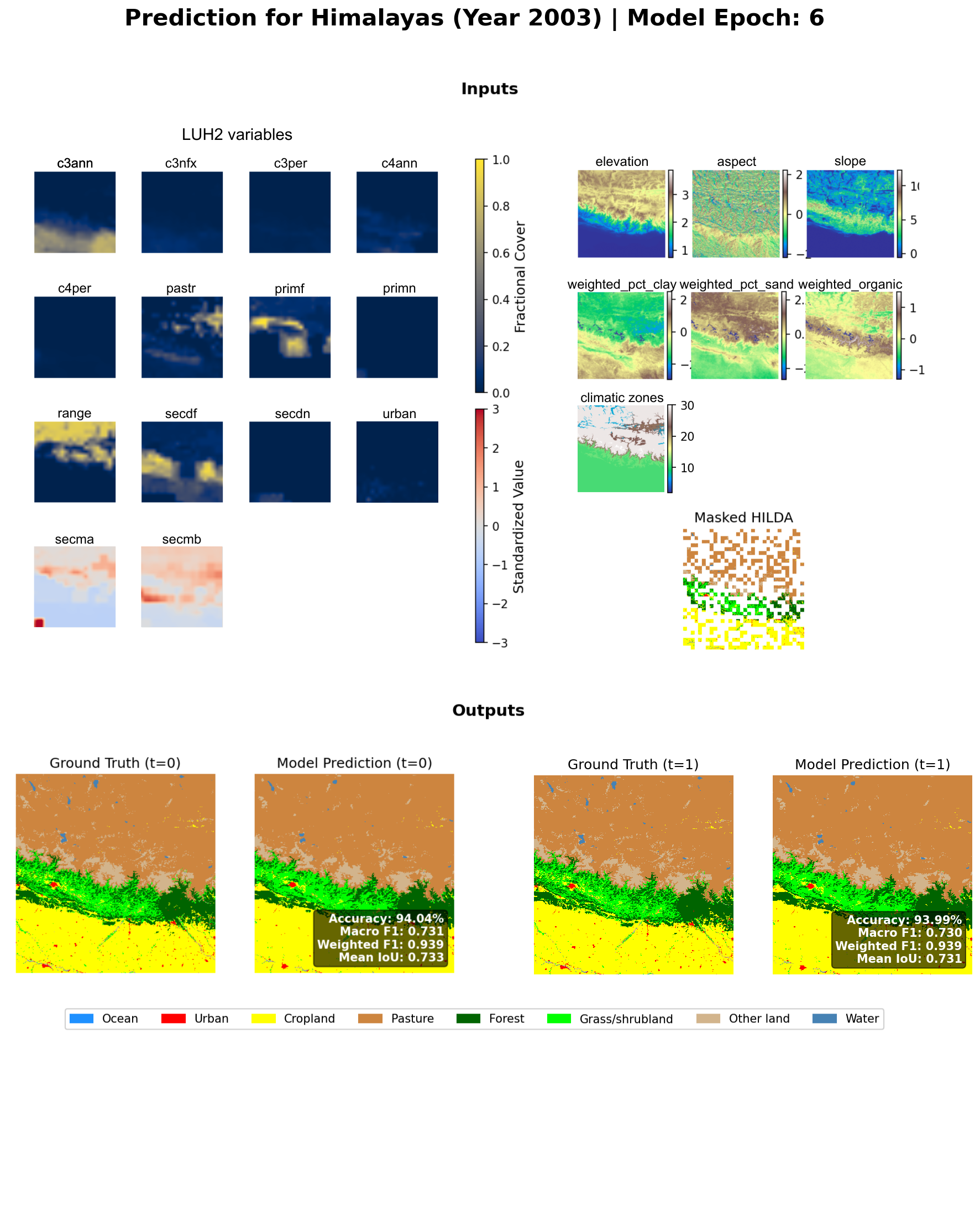}
  \caption{Qualitative results for the Himalayas region
  (Year 2003). The figure displays the full stack of input channels, including 12 coarse fractional LU inputs (top-left), 6
  high-resolution static variables (middle-left), and the masked HILDA prior (bottom-left). The model's multi-step predictions (right) for two time steps ($t=0$, $t=1$) show strong agreement with the ground truth, achieving $\sim$94\% accuracy and a Mean IoU of $\sim$0.73.} \label{fig:qualitative}
\end{figure}

To provide a concrete illustration of the model's output and the
complexities of the classification task, Figure~\ref{fig:qualitative} presents a qualitative example. The figure illustrates the multi-band input channels provided to the U-Net, the ground truth masked prior used for training and evaluation, and the final resulting model predictions. Importantly, these visual examples are provided for two consecutive years, 2004 and 2005, which helps to qualitatively assess the model's consistency and ability to handle temporal variations within the classification task.

\subsection{Global Inference}

\begin{figure}[htbp]
  \centering
  \includegraphics[width=0.90\linewidth]{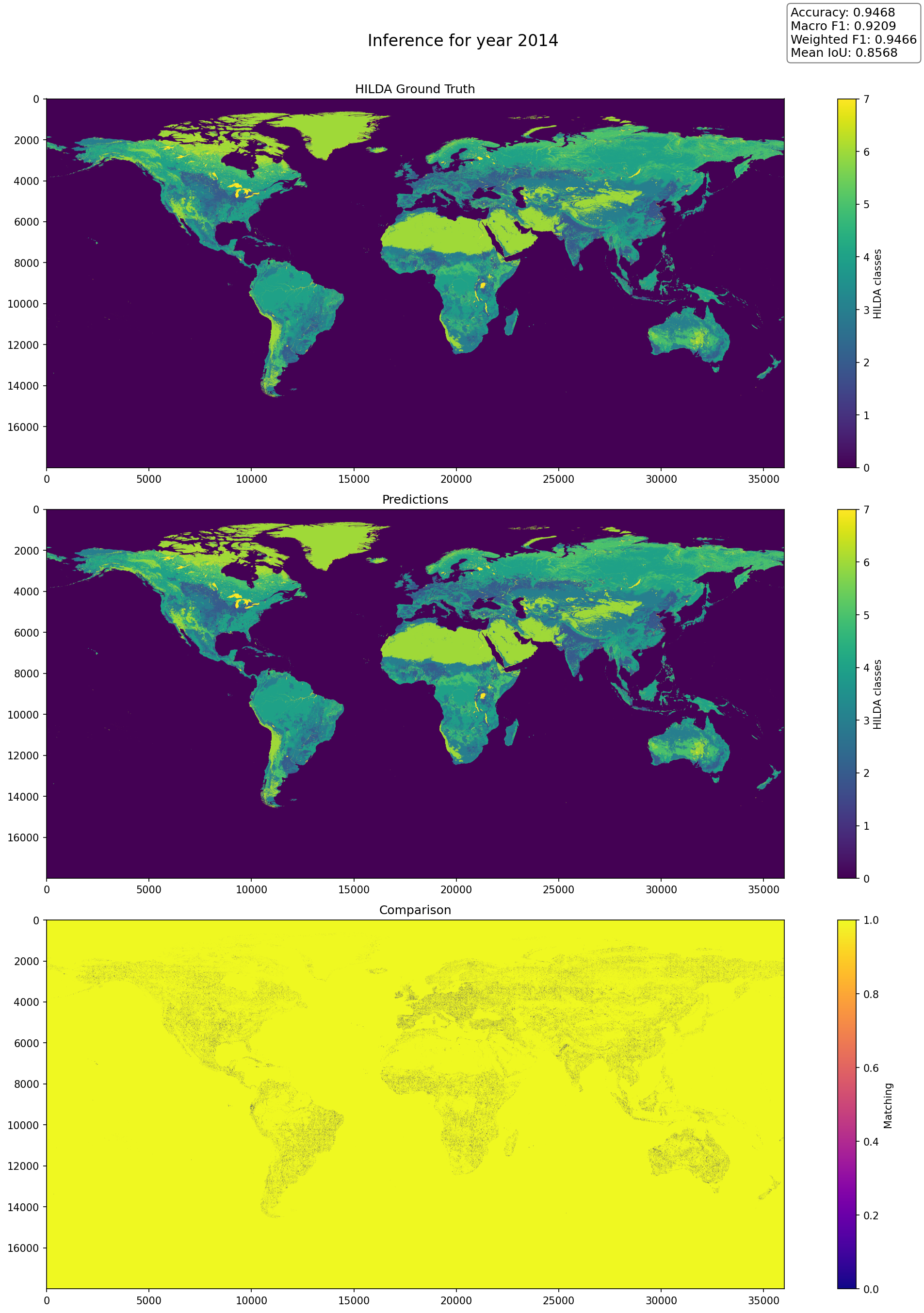}
  \caption{Worldwide inference results (Year 2014). The figure
  shows the ground truth LU data (top), the predictions with our U-Net model (middle), and a heat map of the similarities between these two (bottom).}
  \label{fig:inference}
\end{figure}

The execution of this pipeline produced a comprehensive high-resolution dataset spanning the period 1850--2100, stored in both ARCO Zarr and NetCDF formats. The primary historical output integrates model predictions for the pre-observation era (1850--1899) with consolidated HILDA+ ground truth for the observational record (1899--2020). Furthermore, the system successfully generated three distinct future projection cubes corresponding to the SSP2-4.5, SSP3-7.0, and SSP4-6.0 climate scenarios. Validation of the inference stage, specifically for the year 2014 as shown in Figure~\ref{fig:inference}, demonstrated high fidelity to the ground truth, achieving a global accuracy of 94.88\% and a mIoU of 0.8569, confirming the framework's ability to accurately
reconstruct complex land-use patterns on a global scale.

\section{Discussion and Future Work}

Existing global land-use datasets trade off spatial resolution against temporal coverage: HILDA+ \citep{winkler2020hilda+} offers high resolution but only spans the satellite era, LUH2 \citep{hurtt2020harmonization} covers 850--2100 but at coarse resolution, and Chen et al. \citep{chen2020global} provide global 0.050.05°projections limited to 2015--2100. ML-based downscaling has been demonstrated for specific regions or single countries \citep{yang2020integration, luo20221, truong2024jaxa}, but to our knowledge no model has yet been created that provides a LU dataset bridging this gap at global scale. AI4Land addresses this scaling mismatch by learning the statistical mapping from coarse LUH2 forcing to high-resolution HILDA+ patterns, producing a consistent, high-resolution (1~km) LU dataset covering the historical period (1850--2020) and three future scenarios (up to 2100).

While the underlying U-Net with distributed data-parallel training builds on well-established components, the primary contribution of this work is the demonstration that such a workflow can be deployed on EuroHPC infrastructure to produce a scientifically useful dataset at global scale. The distributed training pipeline achieved parallel efficiency above 97\% up to 8 nodes (32 H100 GPUs) on MareNostrum5; while this represents only a modest fraction of the full system, it indicates that the workflow scales efficiently and provides a foundation for larger-scale runs in future iterations. Qualitative analysis confirms the model's ability to maintain temporal consistency and smooth spatial transitions without edge artifacts. We acknowledge that the LU dataset is inherently impacted by the quality and biases of the source data, specifically the uncertainty in HILDA+ (ground truth) and the forcing data from LUH2. Consequently, our model reflects these inherent biases. While global accuracy is high (94.88\%), performance dropped for underrepresented classes, specifically urban areas.


Several directions are planned to address these limitations and extend the AI4Land framework. First, to improve urban and minority class classification, we are incorporating dynamic annual K\"{o}ppen-Geiger climatic zones (1~km) \citep{beck2023high}  and ISIMIP2b population density data \citep{frieler2017assessing} as additional predictors. These additions are expected to significantly improve the urban class classification by providing richer anthropogenic and climatic context.

Second, to better enforce temporal consistency across  centennial-scale projections, we are integrating a Recurrent U-Net that extends the standard U-Net into a recurrent framework, unrolling the model over multiple time steps. The prediction at step $t$ serves as the ``historic prior'' input for step $t+1$. To address exposure bias during long rollouts, we employ scheduled sampling: training starts with ground-truth priors for stability, then linearly decays to rely on self-predictions, ensuring robustness against accumulated errors over the 1850--2100 projection period. Third, to capture long-range spatial dependencies that CNNs may miss, we plan to investigate generative approaches, specifically Flow Matching U-Nets with Vision Transformer (ViT) backbones. These architectures will enable probabilistic realizations of future land use, quantifying uncertainty and capturing long-range dependencies more effectively than deterministic baselines, providing a robust foundation for Digital Twin initiatives. Building on the results of phase one, the second phase of the framework extends to predict continuous biophysical variables characterizing vegetation state. Specifically, we aim to generate high-resolution maps of LAI$_t$ for a given time step $t$ (monthly or sub-monthly). Finally, we are exploring the possibility of leveraging foundation models — through adapters and fine-tuning — as a replacement for our from-scratch trained models, expecting them to outperform the latter by inheriting capabilities already encoded in the foundation model.

In this study, we introduce the first components of the AI4Land framework for high-resolution, temporally consistent estimation of historical and future land surface boundary conditions essential for weather and climate prediction. We present LU/LC models that generate high-resolution, temporally consistent data with expanded temporal coverage spanning 1850--2100. Future work will focus on incorporating additional predictors to improve performance on under-represented classes. To ensure transparency, reproducibility, and broad accessibility, all datasets, models, code, and pretrained weights will be released as open-source, supporting continued innovation in Earth system modeling.

\section*{Acknowledgements}


This work received funding from the European Union's Horizon Europe Framework Programme through the project CONCERTO (Grant Agreement 101185000), TerraDT (Grant Agreement 101187992) and ELLIOT (Grant Agreement 101214398). AM acknowledges Grant JDC2023-051208-I, funded by MICIU/AEI/10.13039/501100011033. AD and SM acknowledge their AI4S fellowship within the ``Generaci\'{o}n D'' initiative by Red.es. This research was supported by the EuroCC National Competence Centre. We acknowledge EuroHPC Joint Undertaking for awarding project ID EHPC-DEV-2025D03-112 and EHPC-AIF-2025SC01-004 access to MareNostrum5 at BSC, Spain. The authors thank the organizers of the Open Hackathon EuroCC AI Hackathon, especially Milos Maric and Simeon Harrison.

\appendix
\section{Additional Experiments}

\bibliographystyle{unsrtnat}
\bibliography{ref}

\begin{thebibliography}{23}
\providecommand{\natexlab}[1]{#1}
\providecommand{\url}[1]{\texttt{#1}}
\expandafter\ifx\csname urlstyle\endcsname\relax
  \providecommand{\doi}[1]{doi: #1}\else
  \providecommand{\doi}{doi: \begingroup \urlstyle{rm}\Url}\fi

\bibitem[Booth et~al.(2012)Booth, Jones, Collins, Totterdell, Cox, Sitch,
  Huntingford, Betts, Harris, and Lloyd]{booth2012high}
Ben~BB Booth, Chris~D Jones, Mat Collins, Ian~J Totterdell, Peter~M Cox,
  Stephen Sitch, Chris Huntingford, Richard~A Betts, Glen~R Harris, and Jon
  Lloyd.
\newblock High sensitivity of future global warming to land carbon cycle
  processes.
\newblock \emph{Environmental Research Letters}, 7\penalty0 (2):\penalty0
  024002, 2012.

\bibitem[Gier et~al.(2024)Gier, Schlund, Friedlingstein, Jones, Jones, Zaehle,
  and Eyring]{gier2024representation}
Bettina~K Gier, Manuel Schlund, Pierre Friedlingstein, Chris~D Jones, Colin
  Jones, S{\"o}nke Zaehle, and Veronika Eyring.
\newblock Representation of the terrestrial carbon cycle in cmip6.
\newblock \emph{Biogeosciences}, 21\penalty0 (22):\penalty0 5321--5360, 2024.

\bibitem[Zhao et~al.(2010)Zhao, Liu, Li, and Sohl]{zhao2010spatial}
SQ~Zhao, S~Liu, Z~Li, and Terry~L Sohl.
\newblock A spatial resolution threshold of land cover in estimating
  terrestrial carbon sequestration in four counties in georgia and alabama,
  usa.
\newblock \emph{Biogeosciences}, 7\penalty0 (1):\penalty0 71--80, 2010.

\bibitem[Li et~al.(2024)Li, Bisht, Hao, and Leung]{li2024global}
Lingcheng Li, Gautam Bisht, Dalei Hao, and L~Ruby Leung.
\newblock Global 1 km land surface parameters for kilometer-scale earth system
  modeling.
\newblock \emph{Earth System Science Data}, 16\penalty0 (4):\penalty0
  2007--2032, 2024.

\bibitem[Boussetta et~al.(2013)Boussetta, Balsamo, Beljaars, Kral, and
  Jarlan]{boussetta2013impact}
Souhail Boussetta, Gianpaolo Balsamo, Anton Beljaars, Tomas Kral, and Lionel
  Jarlan.
\newblock Impact of a satellite-derived leaf area index monthly climatology in
  a global numerical weather prediction model.
\newblock \emph{International journal of remote sensing}, 34\penalty0
  (9-10):\penalty0 3520--3542, 2013.

\bibitem[Fang et~al.(2019)Fang, Baret, Plummer, and
  Schaepman-Strub]{fang2019overview}
Hongliang Fang, Frederic Baret, Stephen Plummer, and Gabriela Schaepman-Strub.
\newblock An overview of global leaf area index (lai): Methods, products,
  validation, and applications.
\newblock \emph{Reviews of Geophysics}, 57\penalty0 (3):\penalty0 739--799,
  2019.

\bibitem[Prentice et~al.(2024)Prentice, Balzarolo, Bloomfield, Chen, Dechant,
  Ghent, Janssens, Luo, Morfopoulos, Ryu, et~al.]{prentice2024principles}
I~Colin Prentice, Manuela Balzarolo, Keith~J Bloomfield, Jing~M Chen, Benjamin
  Dechant, Darren Ghent, Ivan~A Janssens, Xiangzhong Luo, Catherine
  Morfopoulos, Youngryel Ryu, et~al.
\newblock Principles for satellite monitoring of vegetation carbon uptake.
\newblock \emph{Nature Reviews Earth \& Environment}, 5\penalty0 (11):\penalty0
  818--832, 2024.

\bibitem[Hurtt et~al.(2020)Hurtt, Chini, Sahajpal, Frolking, Bodirsky, Calvin,
  Doelman, Fisk, Fujimori, Goldewijk, et~al.]{hurtt2020harmonization}
George~C Hurtt, Louise Chini, Ritvik Sahajpal, Steve Frolking, Benjamin~L
  Bodirsky, Katherine Calvin, Jonathan~C Doelman, Justin Fisk, Shinichiro
  Fujimori, Kees~Klein Goldewijk, et~al.
\newblock Harmonization of global land-use change and management for the period
  850--2100 (luh2) for cmip6.
\newblock \emph{Geoscientific Model Development Discussions}, 2020:\penalty0
  1--65, 2020.

\bibitem[{Copernicus Global Land Service /
  EEA}(2020)]{Copernicus_LAI_v2_1999_2020}
{Copernicus Global Land Service / EEA}.
\newblock {Leaf Area Index 1999--2020 (raster 1 km), global, 10‑daily} –
  version 2, 2020.
\newblock URL
  \url{https://land.copernicus.eu/en/products/vegetation/leaf-area-index-v2-0-1km}.
\newblock Temporal coverage: 1999--2020; spatial resolution: raster 1 km;
  dekadal (every 10 days).

\bibitem[Winkler et~al.(2020)Winkler, Fuchs, Rounsevell, and
  Herold]{winkler2020hilda+}
Karina Winkler, Richard Fuchs, M~Rounsevell, and Martin Herold.
\newblock Hilda+ global land use change between 1960 and 2019 [dataset].
  pangaea, 2020.

\bibitem[Luo et~al.(2022)Luo, Hu, Chen, Liu, Hou, and Li]{luo20221}
Meng Luo, Guohua Hu, Guangzhao Chen, Xiaojuan Liu, Haiyan Hou, and Xia Li.
\newblock 1 km land use/land cover change of china under comprehensive
  socioeconomic and climate scenarios for 2020--2100.
\newblock \emph{Scientific data}, 9\penalty0 (1):\penalty0 110, 2022.

\bibitem[Lv et~al.(2025)Lv, Gao, Song, Chen, Ye, and Gao]{lv2025land}
Jiaying Lv, Yifan Gao, Changqing Song, Li~Chen, Sijing Ye, and Peichao Gao.
\newblock Land system changes of terrestrial tipping elements on earth under
  global climate pledges: 2000--2100.
\newblock \emph{Scientific Data}, 12\penalty0 (1):\penalty0 163, 2025.

\bibitem[Chen et~al.(2020)Chen, Vernon, Graham, Hejazi, Huang, Cheng, and
  Calvin]{chen2020global}
Min Chen, Chris~R Vernon, Neal~T Graham, Mohamad Hejazi, Maoyi Huang, Yanyan
  Cheng, and Katherine Calvin.
\newblock Global land use for 2015--2100 at 0.05 resolution under diverse
  socioeconomic and climate scenarios.
\newblock \emph{Scientific Data}, 7\penalty0 (1):\penalty0 320, 2020.

\bibitem[Group(2024)]{GEBCO2024}
GEBCO~Compilation Group.
\newblock Gebco 2024 grid.
\newblock Distributed by GEBCO, British Oceanographic Data Centre, 2024.
\newblock URL
  \url{https://www.gebco.net/data-products/gridded-bathymetry-data}.
\newblock Accessed on 27 July 2025.

\bibitem[Schulzweida(2019)]{schulzweida2019cdo}
Uwe Schulzweida.
\newblock Cdo user guide.
\newblock 2019.
\newblock \doi{10.5281/zenodo.3539275}.

\bibitem[Newman(2024)]{newman2024zarr}
D.~J. Newman.
\newblock Zarr storage specification version 2: Cloud-optimized persistence
  using zarr.
\newblock Technical report, NASA Earth Science Data and Information System
  Standards Coordination Office, 2024.
\newblock URL \url{https://doi.org/10.5067/DOC/ESCO/ESDS-RFC-048v1}.

\bibitem[Ronneberger et~al.(2015)Ronneberger, Fischer, and
  Brox]{ronneberger2015u}
Olaf Ronneberger, Philipp Fischer, and Thomas Brox.
\newblock U-net: Convolutional networks for biomedical image segmentation.
\newblock In \emph{Medical Image Computing and Computer-Assisted Intervention
  (MICCAI)}, pages 234--241. Springer, 2015.

\bibitem[Guo and Berkhahn(2016)]{guo2016entityembeddingscategoricalvariables}
Cheng Guo and Felix Berkhahn.
\newblock Entity embeddings of categorical variables, 2016.
\newblock URL \url{https://arxiv.org/abs/1604.06737}.

\bibitem[Gugger et~al.(2022)Gugger, Debut, Wolf, Schmid, Mueller, Mangrulkar,
  Sun, and Bossan]{accelerate}
Sylvain Gugger, Lysandre Debut, Thomas Wolf, Philipp Schmid, Zachary Mueller,
  Sourab Mangrulkar, Marc Sun, and Benjamin Bossan.
\newblock Accelerate: Training and inference at scale made simple, efficient
  and adaptable.
\newblock \url{https://github.com/huggingface/accelerate}, 2022.

\bibitem[Yang et~al.(2020)Yang, Tao, Shi, Ouyang, Pan, Ren, and
  Lu]{yang2020integration}
Jia Yang, Bo~Tao, Hao Shi, Ying Ouyang, Shufen Pan, Wei Ren, and Chaoqun Lu.
\newblock Integration of remote sensing, county-level census, and machine
  learning for century-long regional cropland distribution data reconstruction.
\newblock \emph{International Journal of Applied Earth Observation and
  Geoinformation}, 91:\penalty0 102151, 2020.

\bibitem[Truong et~al.(2024)Truong, Hirayama, Phan, Hoang, Tadono, and
  Nasahara]{truong2024jaxa}
Van~Thinh Truong, Sota Hirayama, Duong~Cao Phan, Thanh~Tung Hoang, Takeo
  Tadono, and Kenlo~Nishida Nasahara.
\newblock Jaxa’s new high-resolution land use land cover map for vietnam
  using a time-feature convolutional neural network.
\newblock \emph{Scientific Reports}, 14\penalty0 (1):\penalty0 3926, 2024.

\bibitem[Beck et~al.(2023)Beck, McVicar, Vergopolan, Berg, Lutsko, Dufour,
  Zeng, Jiang, van Dijk, and Miralles]{beck2023high}
Hylke~E Beck, Tim~R McVicar, Noemi Vergopolan, Alexis Berg, Nicholas~J Lutsko,
  Ambroise Dufour, Zhenzhong Zeng, Xin Jiang, Albert~IJM van Dijk, and Diego~G
  Miralles.
\newblock High-resolution (1 km) k{\"o}ppen-geiger maps for 1901--2099 based on
  constrained cmip6 projections.
\newblock \emph{Scientific data}, 10\penalty0 (1):\penalty0 724, 2023.

\bibitem[Frieler et~al.(2017)Frieler, Lange, Piontek, Reyer, Schewe,
  Warszawski, Zhao, Chini, Denvil, Emanuel, et~al.]{frieler2017assessing}
Katja Frieler, Stefan Lange, Franziska Piontek, Christopher~PO Reyer, Jacob
  Schewe, Lila Warszawski, Fang Zhao, Louise Chini, Sebastien Denvil, Kerry
  Emanuel, et~al.
\newblock Assessing the impacts of 1.5 c global warming--simulation protocol of
  the inter-sectoral impact model intercomparison project (isimip2b).
\newblock \emph{Geoscientific Model Development}, 10\penalty0 (12):\penalty0
  4321--4345, 2017.

\end{thebibliography}

\end{document}